\pdfoutput=1

\documentclass[11pt]{article}

\usepackage[]{acl}

\usepackage{times}
\usepackage{latexsym}

\usepackage[T1]{fontenc}

\usepackage[utf8]{inputenc}

\usepackage{microtype}

\usepackage{inconsolata}

\usepackage{booktabs}
\usepackage{multirow}
\usepackage{graphicx}
\usepackage{xcolor}
\usepackage{soul}
\usepackage{amssymb}
\usepackage{pifont}
\usepackage{pgfplots}
\usepackage{subfigure}

\newcommand{\blue}[1]{\textcolor{blue}{#1}}

\newcommand{\teal}[1]{\textcolor{teal}{#1}}
\newcommand{\lightblue}[1]{\textcolor[RGB]{0,176,240}{#1}}
\newcommand{\lightgreen}[1]{\textcolor[RGB]{0,176,80}{#1}}

\definecolor{greenline}{RGB}{87,156,55}
\definecolor{blueline}{RGB}{66,117,177}
\definecolor{orangeline}{RGB}{234,137,43}

\definecolor{Blueback}{RGB}{218, 227, 243} 
\definecolor{Greenback}{RGB}{226, 240, 217}
\definecolor{Redback}{RGB}{251, 229, 214} 

\newcommand{\blueback}[1]{
  \begingroup
  \sethlcolor{Blueback}
  \textcolor{black}{\hl{#1}}
  \endgroup
}

\newcommand{\redback}[1]{
  \begingroup
  \sethlcolor{Redback}
  \textcolor{black}{\hl{#1}}
  \endgroup
}

\newcommand{\greenback}[1]{
  \begingroup
  \sethlcolor{Greenback}
  \textcolor{black}{\hl{#1}}
  \endgroup
}

\title{Self-Prompting Large Language Models for Zero-Shot Open-Domain QA}

\author{Junlong Li$^{1,4}$ , Jinyuan Wang$^{2,4}$, Zhuosheng Zhang$^{3,*}$,  Hai Zhao$^{1,4,}$\thanks{\ \ Corresponding authors. This paper was partially supported by Joint Research Project of Yangtze River Delta Science and Technology Innovation Community (No. 2022CSJGG1400). } \\ 
$^{1}$Department of Computer Science and Engineering, Shanghai Jiao Tong University\\
$^{2}$SJTU-Paris Elite Institute of Technology, Shanghai Jiao Tong University \\
$^{3}$School of Electronic Information and Electrical Engineering, Shanghai Jiao Tong University\\
$^{4}$Key Laboratory of Shanghai Education Commission for Intelligent Interaction \\
and Cognitive Engineering, Shanghai Jiao Tong University\\
\texttt{\{lockonn, steve\_wang, zhangzs\}@sjtu.edu.cn, zhaohai@cs.sjtu.edu.cn}
}

\begin{document}

\maketitle
\begin{abstract}
Open-Domain Question Answering (ODQA) aims to answer questions without explicitly providing specific background documents. 
This task becomes notably challenging in a zero-shot setting where no data is available to train tailored retrieval-reader models.
While recent Large Language Models (LLMs) like GPT-3 have demonstrated their effectiveness in zero-shot ODQA using direct prompting methods, these methods still fall short of fully harnessing the potential of LLMs when implicitly invoked.
In this paper, we propose a \textbf{Self-Prompting} framework to explicitly utilize the massive knowledge encoded in the parameters of LLMs and their strong instruction understanding abilities. Concretely, we prompt LLMs step by step to generate multiple pseudo QA pairs with background passages and explanations entirely from scratch.
These generated elements are then utilized for in-context learning. 
Experimental results show that our method significantly surpasses previous state-of-the-art zero-shot methods on three widely-used ODQA datasets and even achieves comparable performance with various customized fine-tuned models on full training data. Our code is available at \url{https://github.com/lockon-n/self-prompting}.

\end{abstract}

\section{Introduction}
Open-Domain Question Answering (ODQA) is a longstanding task in natural
language processing that aims to answer questions about world knowledge without explicitly providing specific background documents \citep{voorhees1999trec, huang2020recent,zhu2021retrieving,zhang2022survey}. 
The most common approach for ODQA is to train a Retriever-Reader pipeline \citep{chen-etal-2017-reading}: (i) first retrieving the most related documents of the question, (ii) then applying the reader to extract or generate the final answer conditioned on these documents \citep{karpukhin-etal-2020-dense, lewis2020retrieval, izacard-grave-2021-leveraging}. Despite the decent performance, they rely on large amount of training data and a large external knowledge corpus, so it is hard to expand these methods to zero-shot ODQA where no training data is available. 

With the emergence of Large Langauge Models (LLMs) such as GPT-3 \citep{brown2020language}, OPT \citep{zhang2022opt}, PaLM \citep{chowdhery2022palm}, InstructGPT \citep{ouyang2022training}, researchers start to leverage them for zero-shot ODQA tasks. These LLMs are capable of generating correct answers without any training data and external corpus with direct prompts (like \textit{Q: \{question\} A:}). Some recent works go further to induce the instruction understanding abilities and use the abundant knowledge in their parameters by asking LLMs to generate rationales called \textit{chain-of-thought} \citep{wei2022chain, kojima2022large} or related background information \citep{yu2022generate,sun2022recitation} before answering. 
However, these naive prompting methods for zero-shot ODQA still fail to compete with customized fine-tuned models \citep{yu2022generate}, since they fail to take full advantage of the powerfulness of LLMs.

In this paper, we focus on zero-shot ODQA with no training data and external corpus. Based on the characteristics of this task, we propose \textbf{Self-Prompting} the LLM to explicitly activate different capabilities of LLMs and combine these capabilities to improve performance. In general, we ask LLMs to automatically generate pseudo QA pairs for in-context learning instead of predicting the answers directly as in naive prompting methods. Self-prompting consists of two stages: preparation and inference. In the preparation stage, the LLM is required to generate a pseudo QA dataset in three steps: (i) write a short Wikipedia-style passage, extract named entities in this passage as answers; (ii) raise questions for the answers; (iii) explain each generated QA pair in a short sentence based on the passage. Relying on the strong instruction understanding ability of LLMs, all these sub-tasks can be effectively conducted with simple natural language prompts. In the inference stage, we propose a novel clustering-based retrieval method to select salient examples from this pseudo dataset as the in-context demonstrations for each test sample. These selected QA pairs, along with the passages and explanations, are concatenated with the test question in a specific order as the input sequence, which is fed into the LLM to get the final answer. 

We evaluate our approach on three ODQA benchmarks: WebQ \citep{berant-etal-2013-semantic}, NQ \citep{kwiatkowski-etal-2019-natural}, and TriviaQA \citep{joshi-etal-2017-triviaqa}. Experimental results show that our method significantly surpasses the direct prompting baselines and the previous state-of-the-art (SOTA) zero-shot method \textit{GENREAD} \citep{yu2022generate}. Our method even shows comparable performance with strong few-shot methods. We also conduct extensive analysis on the effects of different generated components, like the formats of the prompted sequence for in-context learning, ways of demonstration selection from the pseudo dataset, and the quality of generated QAs. In general, our contributions can be summarized as follows:

(i) We propose Self-Prompting to leverage multiple capabilities of LLMs for zero-shot ODQA. It can automatically build a pseudo but high-quality ODQA dataset in advance and select salient QA instances for in-context learning. 

(ii) We propose a clustering-based retrieving method to effectively utilize the built pseudo QA dataset to select both semantically similar and diverse examples for each test sample.

(iii) We conduct extensive experiments to show the effectiveness of Self-Prompting on three ODQA tasks. It achieves new SOTA for zero-shot ODQA, and is comparable with some fine-tuned Retriever-Reader models and few-shot prompting methods.

\begin{figure*}
    \centering
    \includegraphics[height=0.45\textwidth]{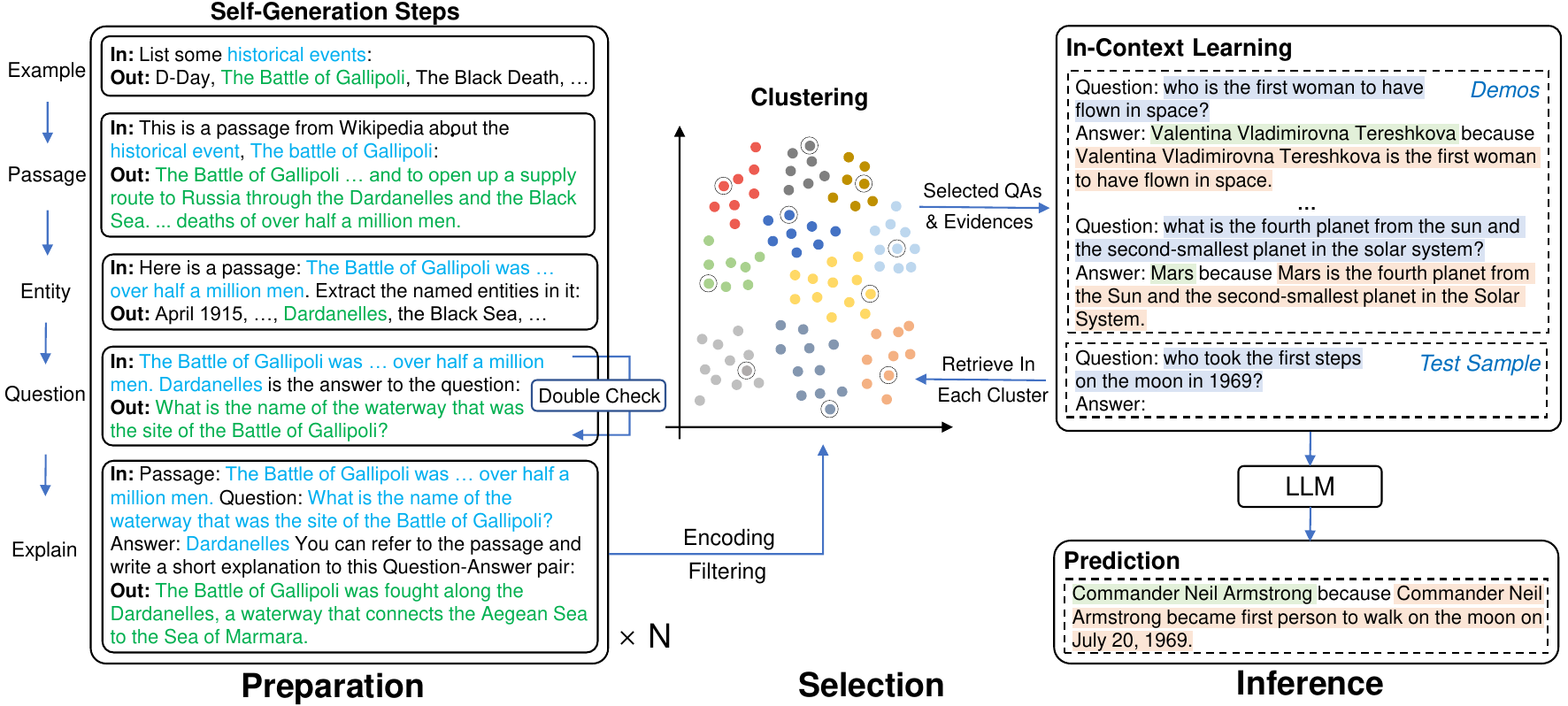}
    \caption{The overall framework for Self-Prompting on ODQA. In the self-generation steps, \lightblue{blue} refers to contents generated in previous steps or manually designed topics, and \lightgreen{green} is the newly generated texts. In inference, \blueback{question}, \greenback{answer}, and \redback{explanation} are question, answer, and explanation for both demonstrations and test samples.}
    \label{fig:overall}
\end{figure*}

\section{Related Works}
\paragraph{Retriever-Reader Models for ODQA}
The mainstream method to tackle ODQA tasks is the Retriever-Reader architecture \citep{zhang-etal-2023-survey-efficient}. It first leverages a retriever over a large knowledge corpus to select several related documents that may contain the answer. Then, a reader is used to process the retrieved documents and predict the answer. Conventional models use sparse retrieval methods such as TF-IDF or BM25, while recent works choose dense retrieval based on the representation vector encoded by pre-trained language models \citep{karpukhin-etal-2020-dense, lewis2020retrieval, guu2020retrieval, izacard-grave-2021-leveraging}. There are also two choices for the reader: extractive readers like BERT \citep{devlin-etal-2019-bert} or generative ones like T5 \citep{raffel2020exploring}. A related work in this branch is PAQ \citep{lewis-etal-2021-paq}, which generates 65 million probably-asked questions based on the complete dump of Wikipedia and directly retrieves these questions instead of documents. 
\paragraph{LLM and In-context Learning}
Generally, LLMs refer to the pre-trained models with tens or hundreds of billions of parameters, such as GPT-3, FLAN, PaLM, OPT, and InstructGPT \citep{brown2020language, wei2022finetuned, chowdhery2022palm, zhang2022opt, ouyang2022training}. These models are trained with large-scale unsupervised learning and are able to perform NLP tasks by converting inputs into natural language queries without further training. The cost of fine-tuning these models is extremely huge, so the usual way of using them is in-context learning, which is to put some input-output pairs as demonstrations in front of the test sample. Some previous works have investigated the calibration \citep{zhao2021calibrate}, example selection \citep{liu-etal-2022-makes,rubin-etal-2022-learning} and ordering \citep{lu-etal-2022-fantastically} of in-context learning, while to the best of our knowledge, we are the first to generate a large number of demonstrations used in in-context learning completely by the LLM itself from scratch.
\paragraph{Enhancing Models with LLM generation}
A recent line of research aims to use the generated contents of LLMs to facilitate the training of small models. For example, \citet{ye2022zerogen} uses GPT-2 \citep{radford2019language} to generate pseudo data to train tiny language models, and \citet{wang2022elaboration} distills the knowledge of GPT-3 into GPT-2 for commonsense QAs. Another line of work directly uses contents generated by the LLM itself. Some works use LLMs to generate relevant contexts or background documents, which are used as additional inputs when answering questions \citep{liu-etal-2022-generated, yu2022generate,sun2022recitation}, while others focus on eliciting a series of intermediate reasoning steps referred to as \textit{chain-of-thought} for arithmetic problems \citep{wei2022chain, kojima2022large, zhang2022automatic}.

\section{Approach}
This section presents the details of Self-Prompting, which has two stages: preparation and inference. In the first stage, we require the LLM to automatically build a pseudo ODQA dataset by prompting it to generate QA pairs with context passages and explanations. In the second stage, we dynamically select examples in the pool through a clustering-based retrieval method as in-context demonstrations to help understand and answer the given question. An overall framework is shown in Figure \ref{fig:overall}.

\subsection{Qseudo QA Dataset Generation}
In the preparation stage, we prompt the LLM to automatically generate a large number of QA pairs as a pseudo dataset by the following four steps. 

\paragraph{Passage Generation}
\label{para:pg}
To ensure the diversity of the generated passages, we first sample various topics (such as \textit{countries, books, tourist attractions}) that are likely to appear in ODQA based on the distribution of hierarchical WordNet synsets for entities appearing in the answers of TriviaQA \citep{joshi-etal-2017-triviaqa}. Specifically, we first abstract and summarize these entities to define a set of high-level themes such as politics, sports, geography, film, television, etc. Then under each high-level theme, we further subdivide it into several topics; for instance, the sports theme is broken down into athletes, sports teams, and sporting events, leading to a complete list of topics. Regarding the number of examples needed for each topic, we empirically set the requirement for each high-level theme at 100 and evenly distribute this value among all the subordinate topics. Subsequently, we manually adjust these numbers, resulting in the data shown in our paper (Table \ref{tab:topic_name_number} in Appendix \ref{app:gen_topic}).

For each topic, the LLM is asked to list some examples with instructions like ``\texttt{List some \{topic\}:}'', and this step is repeated until we have collected a certain number of different examples for it. Through this, we obtain a huge number of examples covering different categories and they are leveraged to generate short Wikipedia-style passages with the following prompt ``\texttt{This is a passage from Wikipedia about the \{topic\}, \{example\}:}''. Note that we do not require the generated article to be an exact copy of the Wikipedia page but just a Wikipedia-style paragraph with enough factual information.

\paragraph{Named Entity Recognition}
We extract the named entities in these generated passages as the candidate answers. Usually, this step is conducted by a fine-tuned small model. In our framework, this is also done by the LLM itself. For a given passage, the prompt might be ``\texttt{Here is a passage: \{passage\} Extract the named entities in it:}'', and we can get the entities in this passage.

\paragraph{Question Generation}
Named entities (e.g., date, name, and location) extracted in the previous step are used as the candidate answers. We then ask the LLM to write a proper question for this answer based on the given passage with prompts like ``\texttt{\{passage\} \{entity\}\ is the answer to the question:}''. To ensure the correctness of the QA pair, we ask the LLM to double-check, i.e., reanswer the question based on the passage to see whether it can recover the entity, and only keep the sample when the answer predicted by the LLM has the same normalized form as the extracted entity.

\paragraph{Explain the QA pair}
For each QA pair, we ask the LLM to return a one-sentence explanation according to the passage. The prompt is like ``\texttt{Passage: \{passage\} Question: \{question\} Answer: \{answer\} You can refer to the passage and write a short explanation to this Question-Answer pair:}''. This step elicits the summarization and induction skills of the LLM to provide a fine-grained annotation for the generated QA pairs without including too much redundant information in the original passage. 

\subsection{Dynamic In-context Demonstrations Selection for Inference}
It is an open question for how to use the pseudo QA dataset generated in the preparation stage. We focus on two aspects, namely selection and format.

\paragraph{Clustering-based Retrieval} Some previous works point out that using examples with high semantic similarity as in-context demonstrations brings benefits \citep{liu-etal-2022-makes}, while others claim that a fixed set of examples based on clustering is better \citep{zhang2022automatic}. We find it effective to combine these two ways. First, each QA pair is encoded into a vector representation over the whole QA pool with Sentence-BERT \citep{reimers-gurevych-2019-sentence}. Suppose $k$ examples are needed for in-context demonstration; the pseudo QAs will be clustered into $k$ categories by the k-means algorithm. For a given question, we encode the question using Sentence-BERT and retrieve the most similar example from each cluster with simple cosine similarity. This selection method balances the similarity and diversity of the demonstrations.

\paragraph{Answer then Explain}
The final step is to organize the format of these selected examples. In the input sequence, we first put these examples sequentially in the format of \textit{Question$\to$ Answer $\to$ Explanation} and place the test question at the end of the sequence.
By doing so, the LLM can view more information than just switching to a QA mode, and it can also give out a brief explanation along with its answer. This is different from the common practice in \textit{chain-of-thought} prompting, i.e., generating a rationale before the answer, but our experiments prove the effectiveness of our choice.

\begin{table*}
  \centering\small
  \caption{Main results on three ODQA benchmarks, Self-Prompting is free from any training data and external knowledge corpus. \# Total Params. is the total number of model parameters in this system (e.g., RAG uses 2 BERT-base with 110M$\times$2 and 1 BART-large with 400M). Train Data refers to whether the system uses training data for training or as in-context demonstrations, and External Corpus means whether an external knowledge corpus is used to retrieve documents. $^\star$This value is obtained by 5-shot in-context learning with samples from the training set as demonstrations. $^\dagger$This value is obtained on a subset of TriviaQA used in RECITE to save computational cost.}
  \setlength{\tabcolsep}{6.5pt}
    \begin{tabular}{l|ccc|cccc}
    \toprule
    \multirow{2}{*}{Models} & \# Total & Train & External & \multirow{2}{*}{WebQ} & \multirow{2}{*}{NQ} & \multirow{2}{*}{TriviaQA} & \multirow{2}{*}{Avg.} \\
& Params. & Data &Corpus \\ 
    \midrule
    \multicolumn{8}{l}{\textit{*fine-tuned models without retrieval}} \\
    T5-SSM \citep{roberts-etal-2020-much}& 11B   & \ding{51} & \ding{55} & 40.8  & 34.8  & 51.0    & 42.2 \\
    \midrule
    \multicolumn{8}{l}{\textit{*retrieval-augmented fine-tuned models}} \\
    REALM \citep{guu2020retrieval}& 330M  & \ding{51} & \ding{51} & 40.7  & 40.4  & 55.8  & 45.6 \\
    DPR \citep{karpukhin-etal-2020-dense} & 330M  & \ding{51} & \ding{51} & 41.1  & 41.5  & 56.8  & 46.5 \\
    RAG \citep{lewis2020retrieval}  & 620M  & \ding{51} & \ding{51} & 45.2  & 44.5  & 56.1  & 48.6 \\
    \midrule
    \multicolumn{8}{l}{\textit{*retrieval-augmented prompting LLMs (DPR trained on target datasets)}} \\
    Google+InstructGPT & 175B  & \ding{55} & \ding{51} & 19.9  & 27.8  & 58.7  & 35.5 \\
    DPR+InstructGPT & 175B  & \ding{55} & \ding{51} & 20.1  & 29.9  & 55.3  & 35.1 \\
    \midrule
    \multicolumn{8}{l}{\textit{*directly prompting LLMs (RECITE is few-shot prompting)}} \\
    InstructGPT & 175B  & \ding{55} & \ding{55} & 18.6  & 20.9  & 52.6  & 30.7 \\
    Codex & 175B  & \ding{55} & \ding{55} & 25.4  & 27.1  & 81.8$^{\dagger\star}$  & -\\
    GENREAD (InstructGPT) \citep{yu2022generate}& 175B  & \ding{55} & \ding{55} & 24.8  & 28.2  & 59.3  & 37.4 \\
    RECITE (Codex) \citep{sun2022recitation}& 175B  & \ding{51} & \ding{55} & -  & 35.8  & 83.5$^\dagger$  & - \\
    
    \midrule
    \multicolumn{8}{l}{\textit{*our method, self prompting by generating pseudo QA for in-context learning}} \\
    Self-Prompting (InstructGPT)  & 175B  & \ding{55} & \ding{55} & 35.6  & 36.2  & 66.8/79.4$^\dagger$  & 46.2 \\
    Self-Prompting (Codex)  & 175B  & \ding{55} & \ding{55} & 38.9  & 40.7  & 84.3$^\dagger$  & - \\
    \bottomrule
    \end{tabular}%
     
  \label{tab:main_res}%
\end{table*}%

\section{Experiments}
\subsection{Datasets and Settings}

We conduct experiments on three English ODQA benchmarks, including WebQ \citep{berant-etal-2013-semantic}, NQ \citep{kwiatkowski-etal-2019-natural} and TriviaQA \citep{joshi-etal-2017-triviaqa}.
Data statistics are in Appendix \ref{app:testset_stat}.

We use InstructGPT \citep{ouyang2022training} (\texttt{text-davinci-002} of GPT-3 \citep{brown2020language}) as the LLM following previous works \citep{wei2022chain,kojima2022large,yu2022generate}. We also conduct experiments on Codex \citep{chen2021evaluating} (\texttt{code-davinci-002} of GPT-3 for code generation) to check the versatility of Self-Prompting. The exact model we used as Sentence-BERT is \texttt{all-mpnet-base-v2}, and the number of demonstrations for in-context learning is 10.

In passage generation, we design 29 topics in advance. Their names and numbers of required examples for each topic are in Appendix \ref{app:gen_topic}. In question generation, we ban some pronouns like \textit{they, he, she} by setting their \texttt{logit\_bias} as -100 in the API call to prevent getting questions that are ambiguous without context (e.g., \textit{what did he do in 1997?}). After the generation process, we filter the QA pair where the answer has more than 5 words, or the LLM outputs an explanation sentence with no target answer span in it. For each passage, the upper limit of generating QA pairs is 10 to ensure diversity. Detailed parameters like \texttt{max\_tokens} or \texttt{temperature} and prompt templates for LLM generation in each step are in Appendix \ref{app:gen_details}. We also provide a brief cost and running efficiency analysis in Appendix \ref{app:cost-running-efficiency}. Following previous works on ODQA, we choose Exact Match (EM) as the evaluation metric, with the same answer normalization as in \citet{karpukhin-etal-2020-dense}. We also observe that in WebQ, if the correct answer contains multiple listed entities, the reference is often given as only one of them. So we heuristically perform extra post-processing on WebQ to extract only the first entity when the LLM predicts multiple ones (e.g., only return \textit{A} if the raw prediction is \textit{A, B, and C}).

The baselines we select include direct prompting: InstructGPT \citep{ouyang2022training}, Codex \citep{chen2021evaluating}, GENREAD \citep{yu2022generate}, RECITE \citep{sun2022recitation}; retrieval-augmented LLM prompting: DPR+InstructGPT, Google+InstructGPT; fine-tuned models with no retrieval: T5-SSM 11B \citep{roberts-etal-2020-much}; retrieval-augmented fine-tined models: REALM \citep{guu2020retrieval}, DPR \citep{karpukhin-etal-2020-dense}, RAG \citep{lewis2020retrieval}.
\subsection{Main Results}
The main results are shown in Table \ref{tab:main_res}. Compared to direct prompting, our Self-Prompting method surpasses the InstructGPT baseline by +15.5 EM on average and the previous SOTA method GENREAD by +8.8 EM on average. A similar improvement is also observed when using Codex as the backbone model, and it even achieves a higher score than RECITE, a strong few-shot baseline. 

Self-Prompting yields comparable results to various powerful retrieval-augmented fine-tuned models with regard to the average EM on three datasets, showing that the LLM itself has stored enough world knowledge in its parameters and provides the potential as an implicit corpus for retrieving. We also notice that Self-Prompting achieves higher EM than T5-SSM 11B on two datasets except WebQ, even though we do not give any training data to the LLM, showing the great potential of LLMs for ODQA under a zero-shot setting. 

Although the generated QA data is not 100\% perfect (see Sec \ref{sec: data_quality}), using them for in-context learning still improves the model performance, which is consistent with the conclusion in \citet{min-etal-2022-rethinking} where they demonstrate that the label space, the distribution of the input text, and the overall format of the sequence are key factors for in-context learning rather than absolute accuracy. Therefore, Self-Prompting can reasonably use the existing capabilities of LLMs to construct data with a similar distribution, and it improves model performance in an in-context learning way as expected.

\section{Analysis}
This section will present our analysis on the construction of in-context learning demonstrations, hyper-parameters in Self-Prompting, the generality of Self-Prompting on different LLMs, and quality analysis of the generated pseudo data. As LLMs like GPT-3 have paywalls, our analysis is conducted on the subsets of the three datasets by randomly selecting 1,000 samples from the test sets. Unless otherwise stated, we use InstructGPT in this section for these three subsets. Besides EM, we also report the F1 score because it is a more robust metric for tiny formal differences between the prediction and gold answer.

\subsection{How to Use Generated Passages and Explanations}
\label{sec:format}
A crucial question is how to use the byproducts, namely the passages and explanations, in a better format for in-context learning. Given a list of 10 demonstrations (Q, A, P, E) $\times$ 10 (by clustering-based retrieval), we investigate several input formats in Table \ref{tab:format_ana}. The \textit{one iteration} methods mean only one call of API is needed to generate the answer and the passage/explanation, while in \textit{two iterations} methods the LLM needs to generate the passage/explanation first and use them in the second API call. Detailed templates for these methods are in Appendix \ref{app:format_template}. 
From Table \ref{tab:format_ana}, we see that the QAE format is the most effective, while the other three answer-then-explain formats (QAP, QAEP, and QAPE) are worse than QAE, which indicates that ntroducing too much redundant information in the form of passage is harmful to the LLM. 
We also find that the two \textit{chain-of-thought} style formats, QEA and QPA, are much worse than other variants. In \citet{lu2022learn} and \citet{wei2022chain}, similar findings are concluded that \textit{chain-of-thought} is beneficial to complex reasoning tasks like math problems but has little impact on commonsense questions. Finally, the \textit{two iterations} methods lead to a significant performance drop, which makes them the worst settings considering the doubled inference time and cost.

\begin{table}
  \centering\small
  \caption{Effects of using different formats for in-context learning in the inference stage. Q = question, A = answer, E = explanation sentence, P = passage. We report the average EM and F1 on three datasets.}
  \setlength{\tabcolsep}{14pt}
    \begin{tabular}{ll|cc}
    \toprule
    Demos & Pred. &EM &F1\\
    \midrule
    \midrule
    \multicolumn{4}{l}{\textit{*one iteration}} \\
    QAE   & Q→AE  & \textbf{48.2} &\textbf{58.3}\\
    QAP   & Q→AP  &46.9 &57.5\\
    QEA   & Q→EA  & 43.3 &53.6 \\
    QPA   & Q→PA  & 40.3 &49.9\\
    QAEP  & Q→AEP & 46.9 &57.3\\
    QAPE  & Q→APE & 46.4 &57.1\\
    \midrule
    \midrule
    \multicolumn{4}{l}{\textit{*two iterations}} \\
    1: QP & Q→P   & -     & -    \\
    2: PQA & PQ→A  & 41.5 &51.8\\
    \midrule
    1: QE & Q→E   & -     & -  \\
    2: EQA & EQ→A  & 43.7 &54.7\\
    \bottomrule
    \end{tabular}%
    
  \label{tab:format_ana}%
\end{table}%

\begin{table}
  \centering\small
  \caption{Different ways for demo selection. We report mean value and standard deviation on 5 runs of Random. The metrics are average EM and F1 on three datasets.}
  \setlength{\tabcolsep}{12pt}
    \begin{tabular}{l|cc}
    \toprule
    Selection &  EM & F1 \\
    \midrule
    Random &$46.4_{\pm1.3}$ &$56.7_{\pm1.0}$\\
    Retrieve & 46.7 &57.1\\
    ClusterCenter & 47.1 &57.2\\
    \midrule
    RetrieveInCluster & \textbf{48.2} &\textbf{58.3}\\
    \bottomrule
    \end{tabular}%
    
  \label{tab:selection_ana}%
\end{table}%

\subsection{Ways of Demonstration Selection}
Since the size of the pseudo QA dataset generated by the LLM is much larger than the number of examples we can put in the input sequence for in-context learning, a proper selection strategy is necessary. We conduct experiments with four settings: randomly selection (Random), retrieving the most similar QAs with cosine similarity globally (Retrieve), selecting the closest QA to the centroid in each cluster (ClusterCenter), and retrieving the most similar QA in each cluster (RetrieveInCluster). The other hyper-parameters are kept the same, i.e., 10 demonstrations and QAE format. Results in Table \ref{tab:selection_ana} show that random selection performs the worst and suffers from instability. The Retrieve and ClusterCenter methods have some improvements over Random.
RetrieveInCluster, proposed to select diverse and semantically similar demonstrations, is robust enough to achieve satisfactory scores among all datasets.

\begin{table}
  \centering\small
  \caption{Performance of Self-Prompting (Codex) on different numbers of generated QA pairs.}
    \begin{tabular}{l|ccccc}
    \toprule
    Data Size & 100   & 500   & 1,000  & 2,000  & full \\
    \midrule
    EM    & 51.8  & 51.7  & 52.2  & 52.4  & \textbf{52.6} \\
    \bottomrule
    \end{tabular}%
  \label{tab:ablation_pseudo_data_size}%
\end{table}%

\begin{figure}
	\centering
	\pgfplotsset{height=5cm,width=7cm,compat=1.12,every axis/.append style={thick},every tick label/.append style={font=\small},legend columns=1 row=2} 

\begin{tikzpicture} \tikzset{every node}=[font=\small] 
\begin{axis} [
                align = center,
                legend cell align={left},
                legend style={at={(0.85,0.625)},anchor=north},
                ymin=45, ymax=60,
                xticklabels={2,4,6,8,10,12,14}, xtick={0,1,2,3,4,5,6},
                xtick distance=13cm,
                ylabel style={align=center},
                xlabel={\# of Demos},
                ylabel={Avg. Score},
                ytick={45,48,51,54,57,60},
                ymajorgrids=true,
                xmajorgrids=true,
                grid style=dashed,
                xtick pos=bottom,
                ytick pos=left,
                ]

\addplot+[color=blueline,
                    mark=triangle,
                    mark size=1.5pt,
                    ] coordinates {(0, 45.63333333
)  (1, 47.16666667
) (2, 47.03333333
) (3, 48
) (4, 48.2
) (5, 47.7
) (6, 48.1
)};
\addlegendentry{\scriptsize EM}

\addplot+ [color=greenline,
                    mark=otimes,
                    mark size=1.5pt,
                    ] coordinates { (0, 55.92
)  (1, 57.23333333
) (2, 57.5
) (3, 57.97
) (4, 58.33333333
) (5, 58.01666667
) (6, 58.35666667
)};
\addlegendentry{\scriptsize F1
}

\end{axis}
\end{tikzpicture}

\caption{Average EM / F1 when using different numbers of demonstrations in Self-Prompting.}
 \label{fig:num_vs_traindata}
\end{figure}
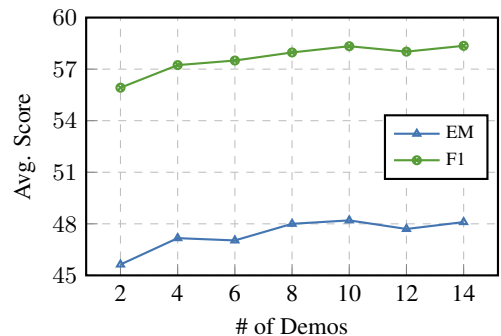

\begin{table*}
  \centering\small
  \caption{The performance of our method for different LLMs. \textbf{Direct} means the models are required to output answers directly with no demonstrations, \textbf{Self-P} (\textbf{Self-Prompting}) uses the self-generated QAs as in-context learning samples.}
  \setlength{\tabcolsep}{5.5pt}
    \begin{tabular}{l|l|l|cccccccccccc}
    \toprule
    Model  & \#Params& Method& \multicolumn{3}{c}{WebQ}  & \multicolumn{3}{c}{NQ}    & \multicolumn{3}{c}{TriviaQA} & \multicolumn{3}{c}{Avg.}\\
    &&&EM&F1&IE&EM&F1&IE&EM&F1&IE&EM&F1&IE\\
    \midrule
    \multirow{2}[2]{*}{InstructGPT} &\multirow{2}[2]{*}{175B}& Direct & 25.3  &38.7& 52.6 & 21.6 &31.9&38.9& 55.1 &64.8&72.0 &34.0&45.1&54.5\\
     && Self-P & 38.7  &51.6&61.4& 37.2&47.6&48.3&68.7 &75.7&78.2&48.2 &58.3 &62.6\\
    \midrule
    \multirow{2}[2]{*}{Codex} &\multirow{2}[2]{*}{175B}& Direct &27.6	&44.5&55.9	&27.1	&38.0&45.8	&67.6	&73.6&76.3	&40.8	&52.1&59.3
\\
     && Self-P &44.0	&55.8&64.6	&41.3	&51.7&51.2	&72.4	&79.5&81.8	&52.6 	&62.3 &65.9\\
    \midrule
    \multirow{2}[2]{*}{GPT-NeoX} &\multirow{2}[2]{*}{20B}& Direct &14.9	&27.7&38.8	&5.9	&12.3&28.7	&19.1	&25.8&37.8	&13.3	&21.9&35.1\\
     && Self-P &27.6	&40.6	&41.7&19.2	&26.5	&26.9&38.3	&43.8&46.4	&28.4 	&37.0 &38.3\\
     \midrule
     \multirow{2}[2]{*}{Alpaca} &\multirow{2}[2]{*}{7B}& Direct &16.4	&32.9	&46.0&11.4	&21.3&29.7	&29.3	&41.5&51.4	&19.0	&31.9&42.4\\
     && Self-P &25.8	&40.3&50.0	&20.3	&30.0&32.6	&44.3	&52.8&57.9	&30.1 	&41.0 &46.8\\
    \bottomrule
    \end{tabular}%
  
  \label{tab:more_models}%
\end{table*}%

\begin{table*}[t]
  \small
  \caption{Two examples in the pseudo QA dataset. 
  \teal{teal}s are the key sentences for the questions and \blue{blue}s are the entities used as answers.}
  \centering
    \begin{tabular}{p{15.5cm}}
    \toprule
    
    \textbf{Topic:} Video game \\ 
    \textbf{Example:} Splatoon \\
    \textbf{Passage:} \teal{Splatoon is a third-person shooter video game developed and published by Nintendo for the} \blue{Wii U}, which was released worldwide in May 2015. The game centers around characters known as Inklings—anthropomorphic squid that can transform between humanoid and cephalopod forms, and use colored ink to attack opponents and clear goals. Splatoon features several game modes, including four-on-four online multiplayer and a single player campaign. \\
    \textbf{Entity:} \blue{the Wii U} \\
    \textbf{Question:} On what platform is Splatoon available? \\
    \textbf{Explanation:} Splatoon is a third-person shooter video game developed and published by Nintendo for \blue{the Wii U}. \\
    \midrule
    \textbf{Topic:} Historical figure \\
    \textbf{Example:} Augustus Caesar \\
    \textbf{Passage:} Augustus Caesar (63 BC – 14 AD) was the first Roman emperor, ruling from 27 BC until his death in 14 AD. He was born Gaius Octavius Thurinus into an old and wealthy equestrian branch of the plebeian gens Octavia. His maternal great-uncle Julius Caesar was assassinated in 44 BC, and Octavius was named in Caesar's will as his adopted son and heir, then known as Octavianus (Anglicized as Octavian). He, Mark Antony, and Marcus Lepidus formed the Second Triumvirate to defeat the assassins of Caesar. Following their victory at Philippi, the Triumvirate divided the Roman Republic among themselves and ruled as military dictators. The Triumvirate was eventually torn apart by the competing ambitions of its members. Lepidus was driven into exile and stripped of his position, \teal{and Antony committed suicide following his defeat at the Battle of Actium by Octavian in} \blue{31 BC}\teal{.}
 \\
    \textbf{Entity:} \blue{31 BC} \\
    \textbf{Question:} In which year did Octavian defeat Antony at the Battle of Actium? \\
    \textbf{Explanation:} The Battle of Actium took place in \blue{31 BC}, and was the decisive battle of the Final War of the Roman Republic. \\
    \bottomrule
    \end{tabular}%
  
  \label{tab:selfgen_examples}%
\end{table*}%

\subsection{Different Number of Demonstrations}
\label{sec:num_analysis}
A natural idea of in-context learning is to put as many examples as possible in the input, so we also investigate the effect of different numbers of demonstrations, as in Figure \ref{fig:num_vs_traindata}. We report the average EM and F1 over the three datasets. When the number ranges from 2 to 10, the performance generally becomes better as the number increases, and using more than 10 examples does not bring significant improvements. As a result, we choose 10 demonstrations in our main experiments for both performance and cost considerations.

\subsection{Impact of the size of generated pseudo data for Self-Prompting}
In our preliminary experiments, we find that Self-Prompting can achieve strong performance with substantially fewer generated examples than our current data size of about 5,000. We report the average EM on the three datasets used for analysis in Table \ref{tab:ablation_pseudo_data_size} for Codex. While we chose the maximum data size for our main experiments to demonstrate the peak capabilities, these results highlight that comparable gains can also be achieved with good data efficiency and greatly reduced overhead.

\begin{figure*}[htbp]
    \centering
    \subfigbottomskip=0pt
    \subfigure[Average \# of flaws in a passage.]{
        \begin{minipage}[t]{0.48\linewidth}
            \centering
            \includegraphics[width=0.95\textwidth]{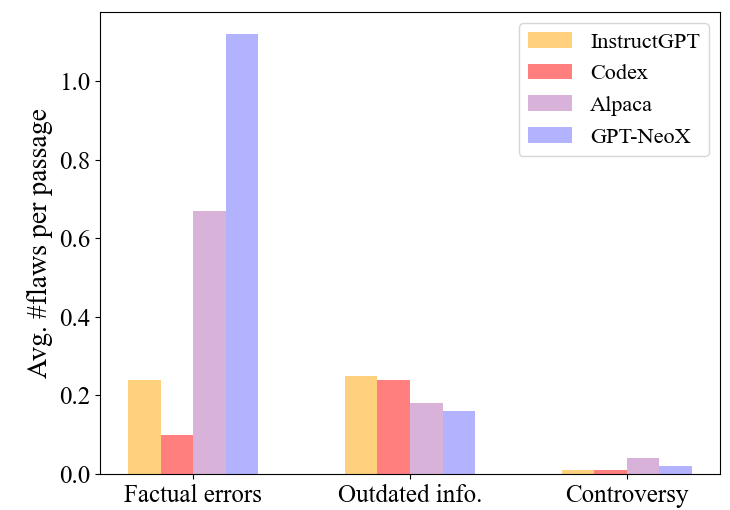}
        \end{minipage}
    }
    \subfigure[Quality of the question, answer and explanation.]{
        \begin{minipage}[t]{0.48\linewidth}
            \centering
            \includegraphics[width=0.95\textwidth]{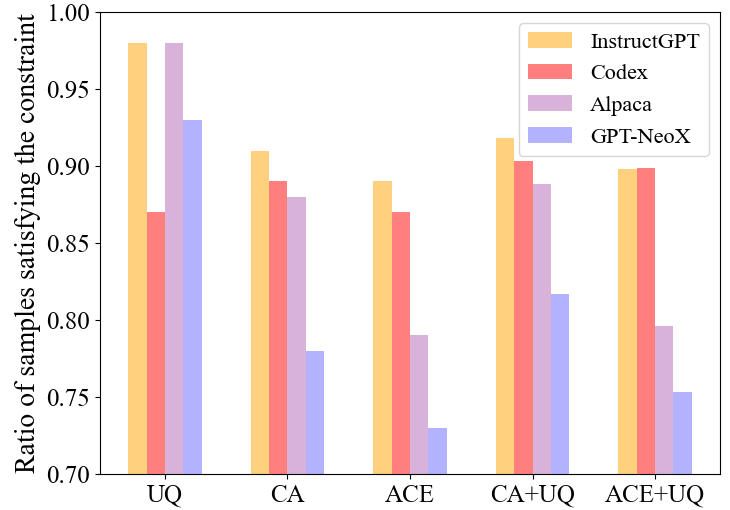}
        \end{minipage}
    }
    \caption{Quantitative study of the generated data. In (b), UQ = Unambiguous question, CA = Correct answer, ACE = Explanation is both accurate and helpful, CA+UQ and ACE+UQ = ratio of CA and ACE given UQ, respectively. In (a) lower value indicates a higher quality, while in (d), a higher value is better.}
    \label{fig:quantitative_eval_data}
\end{figure*}

\subsection{General Effectiveness with LLMs of Different Sizes}
\label{sec:more_models}
Besides InstructGPT and Codex, we also try two smaller LLMs to see how Self-Prompting performs on different sizes of models: GPT-NeoX (20B) \citep{black2022gptneox20b} and Alpaca (7B) \citep{alpaca}. Since the performance of large language models on ODQA is highly under-estimated with EM and F1 \citep{kamalloo-etal-2023-evaluating}, we also report the Instruct-Eval (IE) \citep{kamalloo-etal-2023-evaluating} score with GPT-4 \citep{achiam2023gpt} (\texttt{gpt-4-1106-preview}) as the evaluator model.
The results in Table \ref{tab:more_models} show that for LLMs with parameters of different orders of magnitude, Self-Prompting boosts their performance significantly on all three datasets. It also narrows the performance gap between different LLMs, even if the gap is large under direct prompting. 

\subsection{Data Generation Quality Analysis}
\label{sec: data_quality}
To further explore the quality of the generated pseudo data, we display two examples in Table \ref{tab:selfgen_examples} as a case study, with key sentences in \teal{teal} and answer entities in \blue{blue}. The questions generated for extract entities are proper and answerable with no context given, which is in line with ODQA. As a key part of Self-Prompting, we see that the explanations written by the LLM are reasonable. In the first example, the LLM precisely extracts the key sentence from the passage, while in the second, the LLM not only identifies relevant information but also adds effective explanations. 

We also conduct a quantitative study to detect the hallucination issue \citep{ji2023survey} in the generated contents. We collect the data generated by all four LLMs used in Sec \ref{sec:more_models} and manually selected 100 samples (Q, A, P, E) from each of these four sources with the same topics (e.g., passages and QAs generated by different models but all about \textit{The Legend of Zelda}) for fair comparison. For each sample, we use GPT-4 \citep{openai2023gpt4} based New Bing\footnote{https://bing.com/chat} to check: (i) the number of factual errors, outdated information and controversial opinions in the generated passage; (ii) if the question is unambiguous without context; (iii) if the answer is correct for the question; (iv) if the explanation is both accurate and helpful for question answering. 

All LLMs inevitably generate outdated information, as they cannot access the Internet and are limited by training data. The two smaller models, Alpaca and GPT-NeoX, make significantly more factual mistakes when they generate the required passages, as shown in Figure \ref{fig:quantitative_eval_data}(a) while InstructGPT and Codex only have one factual error in every 5 and 10 passages respectively. 
From Figure \ref{fig:quantitative_eval_data}(b), we see that (question, answer, explanation) triples by InstructGPT are better than those from other models on almost all concerned metrics, showing its strong instruction understanding ability.

\begin{table}
  \centering\small
  \caption{Comparison between Self-Prompting and in-context learning with training data (Traindata-ICL). We report average EM/F1 on the non-overlapping subsets.}
  \setlength{\tabcolsep}{12pt}
    \begin{tabular}{l|cc}
    \toprule
    Method & EM & F1 \\
    \midrule
    Traindata-ICL & 36.0     & 48.7 \\
    Self-Prompting & 34.2     & 46.0 \\
    \bottomrule
    \end{tabular}%
  \label{tab:selfp_vs_traindataicl}%
\end{table}%

\subsection{Comparison between Self-Prompting and Using Training Data}
\label{sec:selfp_vs_using-training-data}
Since the generated pseudo data is not flawless, it is natural to see the gap between Self-Prompting and in-context learning with real training data as a strong baseline (Traindata-ICL). To avoid the test-train data overlap problem, we use the non-overlapping subsets annotated in \citet{lewis-etal-2021-question} (425 WebQ samples, 357 NQ samples, and 254 TriviaQA samples) to conduct experiments in this section. Since there are no explanations in the training data, we only use a simple \textit{question-answer} format. Other procedures and hyper-parameters, like the clustering-based selection and 10 in-context demonstrations, are kept the same. We report the average EM and F1 on the three datasets. 
The results in Table \ref{tab:selfp_vs_traindataicl} show that even though the size of available demonstrations is much larger (e.g., 70K training data for NQ v.s. 4.8K pseudo data by Self-Prompting), Self-Prompting still obtains comparable performance with using training data for in-context learning (performance gap less than 2 EM and 3 F1). This indicates that even without laborious and costly human annotation, the LLMs can still effectively elicit their own knowledge and answer open-domain questions.

\section{Conclusion}
In this paper, we propose Self-Prompting LLMs for Zero-Shot Open-Domain Question Answering. Our method induces the LLM to generate pseudo QA pairs with matching passages and explanations and use them for in-context learning. It successfully stimulates the potential of the LLM in ODQA by explicitly activating various language understanding abilities and eliciting knowledge in its parameters. With no training data and external corpus, Self-Prompting surpasses previous SOTA significantly and performs on par with several strong retrieval-augmented fine-tuned models and few-shot prompting methods. It is generally effective across LLMs under different sizes and is able to provide accurate and helpful explanations alongside the answers, making it a more reliable and interpretable framework for turstworthy AI systems.


\section{Limitations}
In this paper, various experiments rely on the OpenAI APIs, whose usage needs to be paid and may cause a huge cost when generating a large number of tokens. Therefore, this may affect the reproducibility of this work. We also note that the pseudo QA generating process needs some trial and error to find good prompt templates for each step as shown in Table \ref{tab:gen_details} in Appendix \ref{app:gen_details}, so there is still room to reduce the human effort in the whole process.

\section{Ethics}
In this work, we use three ODQA datasets for experiments. The NQ and TriviaQA datasets use the Apache-2.0 code, and the license for WebQ is not specified. 
In Self-Prompting, any suggested prompt templates refrain from gathering or employing personal information concerning other individuals. The specific prompt templates employed can be found in Appendix \ref{app:gen_details} and \ref{app:format_template}. Notably, all prompt templates in this paper are devoid of any discriminatory language targeting individuals or groups, and they do not pose any potential harm to the safety of others.

We also recognize the importance of considering the potential societal impacts of the synthesized data. To address this, we have primarily utilized the OpenAI Moderation API\footnote{https://platform.openai.com/docs/guides/moderation} to detect if Self-Prompting generated content is free from harmful material. We find that all 977 passages and 4,883 question-answer-explanation triples revealed no toxic content. While our current synthetic data is free from toxicity, we acknowledge the need for safeguards against potential risks in larger-scale generation, especially for contents that are not toxic but contain explicit or implicit bias.  Introducing measures such as the OpenAI Moderation API to detect and halt toxic outputs, setting rules in the prompts for LLM behavior \citep{bai2022constitutional}, and employing controllable decoding strategies, from token bans to sophisticated methods that dynamically shift the output token distribution \citep{li2023rain, liu2021dexperts}, are possible solutions to manage this issue.

\bibliography{anthology,custom}

\begin{thebibliography}{49}
\expandafter\ifx\csname natexlab\endcsname\relax\def\natexlab#1{#1}\fi

\bibitem[{Bai et~al.(2022)Bai, Kadavath, Kundu, Askell, Kernion, Jones, Chen, Goldie, Mirhoseini, McKinnon et~al.}]{bai2022constitutional}
Yuntao Bai, Saurav Kadavath, Sandipan Kundu, Amanda Askell, Jackson Kernion, Andy Jones, Anna Chen, Anna Goldie, Azalia Mirhoseini, Cameron McKinnon, et~al. 2022.
\newblock Constitutional ai: Harmlessness from ai feedback.
\newblock \emph{arXiv preprint arXiv:2212.08073}.

\bibitem[{Berant et~al.(2013)Berant, Chou, Frostig, and Liang}]{berant-etal-2013-semantic}
Jonathan Berant, Andrew Chou, Roy Frostig, and Percy Liang. 2013.
\newblock \href {https://aclanthology.org/D13-1160} {Semantic parsing on {F}reebase from question-answer pairs}.
\newblock In \emph{Proceedings of the 2013 Conference on Empirical Methods in Natural Language Processing}, pages 1533--1544, Seattle, Washington, USA. Association for Computational Linguistics.

\bibitem[{Black et~al.(2022)Black, Biderman, Hallahan, Anthony, Gao, Golding, He, Leahy, McDonell, Phang, Pieler, Prashanth, Purohit, Reynolds, Tow, Wang, and Weinbach}]{black2022gptneox20b}
Sid Black, Stella Biderman, Eric Hallahan, Quentin Anthony, Leo Gao, Laurence Golding, Horace He, Connor Leahy, Kyle McDonell, Jason Phang, Michael Pieler, USVSN~Sai Prashanth, Shivanshu Purohit, Laria Reynolds, Jonathan Tow, Ben Wang, and Samuel Weinbach. 2022.
\newblock \href {https://arxiv.org/abs/2204.06745} {Gpt-neox-20b: An open-source autoregressive language model}.
\newblock \emph{arXiv preprint arXiv:2204.06745}.

\bibitem[{Brown et~al.(2020)Brown, Mann, Ryder, Subbiah, Kaplan, Dhariwal, Neelakantan, Shyam, Sastry, Askell, Agarwal, Herbert-Voss, Krueger, Henighan, Child, Ramesh, Ziegler, Wu, Winter, Hesse, Chen, Sigler, Litwin, Gray, Chess, Clark, Berner, McCandlish, Radford, Sutskever, and Amodei}]{brown2020language}
Tom Brown, Benjamin Mann, Nick Ryder, Melanie Subbiah, Jared~D Kaplan, Prafulla Dhariwal, Arvind Neelakantan, Pranav Shyam, Girish Sastry, Amanda Askell, Sandhini Agarwal, Ariel Herbert-Voss, Gretchen Krueger, Tom Henighan, Rewon Child, Aditya Ramesh, Daniel Ziegler, Jeffrey Wu, Clemens Winter, Chris Hesse, Mark Chen, Eric Sigler, Mateusz Litwin, Scott Gray, Benjamin Chess, Jack Clark, Christopher Berner, Sam McCandlish, Alec Radford, Ilya Sutskever, and Dario Amodei. 2020.
\newblock \href {https://proceedings.neurips.cc/paper/2020/file/1457c0d6bfcb4967418bfb8ac142f64a-Paper.pdf} {Language models are few-shot learners}.
\newblock In \emph{Advances in Neural Information Processing Systems}, volume~33, pages 1877--1901. Curran Associates, Inc.

\bibitem[{Chen et~al.(2017)Chen, Fisch, Weston, and Bordes}]{chen-etal-2017-reading}
Danqi Chen, Adam Fisch, Jason Weston, and Antoine Bordes. 2017.
\newblock \href {https://doi.org/10.18653/v1/P17-1171} {Reading {W}ikipedia to answer open-domain questions}.
\newblock In \emph{Proceedings of the 55th Annual Meeting of the Association for Computational Linguistics (Volume 1: Long Papers)}, pages 1870--1879, Vancouver, Canada. Association for Computational Linguistics.

\bibitem[{Chen et~al.(2021)Chen, Tworek, Jun, Yuan, Pinto, Kaplan, Edwards, Burda, Joseph, Brockman et~al.}]{chen2021evaluating}
Mark Chen, Jerry Tworek, Heewoo Jun, Qiming Yuan, Henrique Ponde de~Oliveira Pinto, Jared Kaplan, Harri Edwards, Yuri Burda, Nicholas Joseph, Greg Brockman, et~al. 2021.
\newblock \href {https://arxiv.org/abs/2107.03374} {Evaluating large language models trained on code}.
\newblock \emph{arXiv preprint arXiv:2107.03374}.

\bibitem[{Chowdhery et~al.(2022)Chowdhery, Narang, Devlin, Bosma, Mishra, Roberts, Barham, Chung, Sutton, Gehrmann, Schuh, Shi, Tsvyashchenko, Maynez, Rao, Barnes, Tay, Shazeer, Prabhakaran, Reif, Du, Hutchinson, Pope, Bradbury, Austin, Isard, Gur-Ari, Yin, Duke, Levskaya, Ghemawat, Dev, Michalewski, Garcia, Misra, Robinson, Fedus, Zhou, Ippolito, Luan, Lim, Zoph, Spiridonov, Sepassi, Dohan, Agrawal, Omernick, Dai, Pillai, Pellat, Lewkowycz, Moreira, Child, Polozov, Lee, Zhou, Wang, Saeta, Diaz, Firat, Catasta, Wei, Meier-Hellstern, Eck, Dean, Petrov, and Fiedel}]{chowdhery2022palm}
Aakanksha Chowdhery, Sharan Narang, Jacob Devlin, Maarten Bosma, Gaurav Mishra, Adam Roberts, Paul Barham, Hyung~Won Chung, Charles Sutton, Sebastian Gehrmann, Parker Schuh, Kensen Shi, Sasha Tsvyashchenko, Joshua Maynez, Abhishek Rao, Parker Barnes, Yi~Tay, Noam Shazeer, Vinodkumar Prabhakaran, Emily Reif, Nan Du, Ben Hutchinson, Reiner Pope, James Bradbury, Jacob Austin, Michael Isard, Guy Gur-Ari, Pengcheng Yin, Toju Duke, Anselm Levskaya, Sanjay Ghemawat, Sunipa Dev, Henryk Michalewski, Xavier Garcia, Vedant Misra, Kevin Robinson, Liam Fedus, Denny Zhou, Daphne Ippolito, David Luan, Hyeontaek Lim, Barret Zoph, Alexander Spiridonov, Ryan Sepassi, David Dohan, Shivani Agrawal, Mark Omernick, Andrew~M. Dai, Thanumalayan~Sankaranarayana Pillai, Marie Pellat, Aitor Lewkowycz, Erica Moreira, Rewon Child, Oleksandr Polozov, Katherine Lee, Zongwei Zhou, Xuezhi Wang, Brennan Saeta, Mark Diaz, Orhan Firat, Michele Catasta, Jason Wei, Kathy Meier-Hellstern, Douglas Eck, Jeff Dean, Slav Petrov, and Noah Fiedel. 2022.
\newblock \href {https://arxiv.org/abs/2204.02311} {Palm: Scaling language modeling with pathways}.
\newblock \emph{arXiv preprint arXiv:2204.02311}.

\bibitem[{Devlin et~al.(2019)Devlin, Chang, Lee, and Toutanova}]{devlin-etal-2019-bert}
Jacob Devlin, Ming-Wei Chang, Kenton Lee, and Kristina Toutanova. 2019.
\newblock \href {https://doi.org/10.18653/v1/N19-1423} {{BERT}: Pre-training of deep bidirectional transformers for language understanding}.
\newblock In \emph{Proceedings of the 2019 Conference of the North {A}merican Chapter of the Association for Computational Linguistics: Human Language Technologies, Volume 1 (Long and Short Papers)}, pages 4171--4186, Minneapolis, Minnesota. Association for Computational Linguistics.

\bibitem[{Guu et~al.(2020)Guu, Lee, Tung, Pasupat, and Chang}]{guu2020retrieval}
Kelvin Guu, Kenton Lee, Zora Tung, Panupong Pasupat, and Mingwei Chang. 2020.
\newblock \href {http://proceedings.mlr.press/v119/guu20a.html?ref=https://githubhelp.com} {Retrieval augmented language model pre-training}.
\newblock In \emph{International Conference on Machine Learning}, pages 3929--3938. PMLR.

\bibitem[{Huang et~al.(2020)Huang, Xu, Hu, Wang, Qiu, Fu, Zhao, Peng, and Wang}]{huang2020recent}
Zhen Huang, Shiyi Xu, Minghao Hu, Xinyi Wang, Jinyan Qiu, Yongquan Fu, Yuncai Zhao, Yuxing Peng, and Changjian Wang. 2020.
\newblock \href {https://ieeexplore.ieee.org/document/9072442} {Recent trends in deep learning based open-domain textual question answering systems}.
\newblock \emph{IEEE Access}, 8:94341--94356.

\bibitem[{Izacard and Grave(2021)}]{izacard-grave-2021-leveraging}
Gautier Izacard and Edouard Grave. 2021.
\newblock \href {https://doi.org/10.18653/v1/2021.eacl-main.74} {Leveraging passage retrieval with generative models for open domain question answering}.
\newblock In \emph{Proceedings of the 16th Conference of the European Chapter of the Association for Computational Linguistics: Main Volume}, pages 874--880, Online. Association for Computational Linguistics.

\bibitem[{Ji et~al.(2023)Ji, Lee, Frieske, Yu, Su, Xu, Ishii, Bang, Madotto, and Fung}]{ji2023survey}
Ziwei Ji, Nayeon Lee, Rita Frieske, Tiezheng Yu, Dan Su, Yan Xu, Etsuko Ishii, Ye~Jin Bang, Andrea Madotto, and Pascale Fung. 2023.
\newblock Survey of hallucination in natural language generation.
\newblock \emph{ACM Computing Surveys}, 55(12):1--38.

\bibitem[{Joshi et~al.(2017)Joshi, Choi, Weld, and Zettlemoyer}]{joshi-etal-2017-triviaqa}
Mandar Joshi, Eunsol Choi, Daniel Weld, and Luke Zettlemoyer. 2017.
\newblock \href {https://doi.org/10.18653/v1/P17-1147} {{T}rivia{QA}: A large scale distantly supervised challenge dataset for reading comprehension}.
\newblock In \emph{Proceedings of the 55th Annual Meeting of the Association for Computational Linguistics (Volume 1: Long Papers)}, pages 1601--1611, Vancouver, Canada. Association for Computational Linguistics.

\bibitem[{Kamalloo et~al.(2023)Kamalloo, Dziri, Clarke, and Rafiei}]{kamalloo-etal-2023-evaluating}
Ehsan Kamalloo, Nouha Dziri, Charles Clarke, and Davood Rafiei. 2023.
\newblock \href {https://doi.org/10.18653/v1/2023.acl-long.307} {Evaluating open-domain question answering in the era of large language models}.
\newblock In \emph{Proceedings of the 61st Annual Meeting of the Association for Computational Linguistics (Volume 1: Long Papers)}, pages 5591--5606, Toronto, Canada. Association for Computational Linguistics.

\bibitem[{Karpukhin et~al.(2020)Karpukhin, Oguz, Min, Lewis, Wu, Edunov, Chen, and Yih}]{karpukhin-etal-2020-dense}
Vladimir Karpukhin, Barlas Oguz, Sewon Min, Patrick Lewis, Ledell Wu, Sergey Edunov, Danqi Chen, and Wen-tau Yih. 2020.
\newblock \href {https://doi.org/10.18653/v1/2020.emnlp-main.550} {Dense passage retrieval for open-domain question answering}.
\newblock In \emph{Proceedings of the 2020 Conference on Empirical Methods in Natural Language Processing (EMNLP)}, pages 6769--6781, Online. Association for Computational Linguistics.

\bibitem[{Kojima et~al.(2022)Kojima, Gu, Reid, Matsuo, and Iwasawa}]{kojima2022large}
Takeshi Kojima, Shixiang~Shane Gu, Machel Reid, Yutaka Matsuo, and Yusuke Iwasawa. 2022.
\newblock \href {https://arxiv.org/abs/2205.11916} {Large language models are zero-shot reasoners}.
\newblock \emph{arXiv preprint arXiv:2205.11916}.

\bibitem[{Kwiatkowski et~al.(2019)Kwiatkowski, Palomaki, Redfield, Collins, Parikh, Alberti, Epstein, Polosukhin, Devlin, Lee, Toutanova, Jones, Kelcey, Chang, Dai, Uszkoreit, Le, and Petrov}]{kwiatkowski-etal-2019-natural}
Tom Kwiatkowski, Jennimaria Palomaki, Olivia Redfield, Michael Collins, Ankur Parikh, Chris Alberti, Danielle Epstein, Illia Polosukhin, Jacob Devlin, Kenton Lee, Kristina Toutanova, Llion Jones, Matthew Kelcey, Ming-Wei Chang, Andrew~M. Dai, Jakob Uszkoreit, Quoc Le, and Slav Petrov. 2019.
\newblock \href {https://doi.org/10.1162/tacl_a_00276} {Natural questions: A benchmark for question answering research}.
\newblock \emph{Transactions of the Association for Computational Linguistics}, 7:452--466.

\bibitem[{Lewis et~al.(2020)Lewis, Perez, Piktus, Petroni, Karpukhin, Goyal, K{\"u}ttler, Lewis, Yih, Rockt{\"a}schel et~al.}]{lewis2020retrieval}
Patrick Lewis, Ethan Perez, Aleksandra Piktus, Fabio Petroni, Vladimir Karpukhin, Naman Goyal, Heinrich K{\"u}ttler, Mike Lewis, Wen-tau Yih, Tim Rockt{\"a}schel, et~al. 2020.
\newblock \href {https://proceedings.neurips.cc/paper/2020/hash/6b493230205f780e1bc26945df7481e5-Abstract.html} {Retrieval-augmented generation for knowledge-intensive nlp tasks}.
\newblock \emph{Advances in Neural Information Processing Systems}, 33:9459--9474.

\bibitem[{Lewis et~al.(2021{\natexlab{a}})Lewis, Stenetorp, and Riedel}]{lewis-etal-2021-question}
Patrick Lewis, Pontus Stenetorp, and Sebastian Riedel. 2021{\natexlab{a}}.
\newblock \href {https://doi.org/10.18653/v1/2021.eacl-main.86} {Question and answer test-train overlap in open-domain question answering datasets}.
\newblock In \emph{Proceedings of the 16th Conference of the European Chapter of the Association for Computational Linguistics: Main Volume}, pages 1000--1008, Online. Association for Computational Linguistics.

\bibitem[{Lewis et~al.(2021{\natexlab{b}})Lewis, Wu, Liu, Minervini, K{\"u}ttler, Piktus, Stenetorp, and Riedel}]{lewis-etal-2021-paq}
Patrick Lewis, Yuxiang Wu, Linqing Liu, Pasquale Minervini, Heinrich K{\"u}ttler, Aleksandra Piktus, Pontus Stenetorp, and Sebastian Riedel. 2021{\natexlab{b}}.
\newblock \href {https://doi.org/10.1162/tacl_a_00415} {{PAQ}: 65 million probably-asked questions and what you can do with them}.
\newblock \emph{Transactions of the Association for Computational Linguistics}, 9:1098--1115.

\bibitem[{Li et~al.(2023)Li, Wei, Zhao, Zhang, and Zhang}]{li2023rain}
Yuhui Li, Fangyun Wei, Jinjing Zhao, Chao Zhang, and Hongyang Zhang. 2023.
\newblock Rain: Your language models can align themselves without finetuning.
\newblock \emph{arXiv preprint arXiv:2309.07124}.

\bibitem[{Liu et~al.(2021)Liu, Sap, Lu, Swayamdipta, Bhagavatula, Smith, and Choi}]{liu2021dexperts}
Alisa Liu, Maarten Sap, Ximing Lu, Swabha Swayamdipta, Chandra Bhagavatula, Noah~A Smith, and Yejin Choi. 2021.
\newblock Dexperts: Decoding-time controlled text generation with experts and anti-experts.
\newblock In \emph{Proceedings of the 59th Annual Meeting of the Association for Computational Linguistics and the 11th International Joint Conference on Natural Language Processing (Volume 1: Long Papers)}, pages 6691--6706.

\bibitem[{Liu et~al.(2022{\natexlab{a}})Liu, Shen, Zhang, Dolan, Carin, and Chen}]{liu-etal-2022-makes}
Jiachang Liu, Dinghan Shen, Yizhe Zhang, Bill Dolan, Lawrence Carin, and Weizhu Chen. 2022{\natexlab{a}}.
\newblock \href {https://doi.org/10.18653/v1/2022.deelio-1.10} {What makes good in-context examples for {GPT}-3?}
\newblock In \emph{Proceedings of Deep Learning Inside Out (DeeLIO 2022): The 3rd Workshop on Knowledge Extraction and Integration for Deep Learning Architectures}, pages 100--114, Dublin, Ireland and Online. Association for Computational Linguistics.

\bibitem[{Liu et~al.(2022{\natexlab{b}})Liu, Liu, Lu, Welleck, West, Le~Bras, Choi, and Hajishirzi}]{liu-etal-2022-generated}
Jiacheng Liu, Alisa Liu, Ximing Lu, Sean Welleck, Peter West, Ronan Le~Bras, Yejin Choi, and Hannaneh Hajishirzi. 2022{\natexlab{b}}.
\newblock \href {https://doi.org/10.18653/v1/2022.acl-long.225} {Generated knowledge prompting for commonsense reasoning}.
\newblock In \emph{Proceedings of the 60th Annual Meeting of the Association for Computational Linguistics (Volume 1: Long Papers)}, pages 3154--3169, Dublin, Ireland. Association for Computational Linguistics.

\bibitem[{Lu et~al.(2022{\natexlab{a}})Lu, Mishra, Xia, Qiu, Chang, Zhu, Tafjord, Clark, and Kalyan}]{lu2022learn}
Pan Lu, Swaroop Mishra, Tony Xia, Liang Qiu, Kai-Wei Chang, Song-Chun Zhu, Oyvind Tafjord, Peter Clark, and Ashwin Kalyan. 2022{\natexlab{a}}.
\newblock \href {https://arxiv.org/abs/2209.09513v2} {Learn to explain: Multimodal reasoning via thought chains for science question answering}.
\newblock In \emph{The 36th Conference on Neural Information Processing Systems (NeurIPS)}.

\bibitem[{Lu et~al.(2022{\natexlab{b}})Lu, Bartolo, Moore, Riedel, and Stenetorp}]{lu-etal-2022-fantastically}
Yao Lu, Max Bartolo, Alastair Moore, Sebastian Riedel, and Pontus Stenetorp. 2022{\natexlab{b}}.
\newblock \href {https://doi.org/10.18653/v1/2022.acl-long.556} {Fantastically ordered prompts and where to find them: Overcoming few-shot prompt order sensitivity}.
\newblock In \emph{Proceedings of the 60th Annual Meeting of the Association for Computational Linguistics (Volume 1: Long Papers)}, pages 8086--8098, Dublin, Ireland. Association for Computational Linguistics.

\bibitem[{Min et~al.(2022)Min, Lyu, Holtzman, Artetxe, Lewis, Hajishirzi, and Zettlemoyer}]{min-etal-2022-rethinking}
Sewon Min, Xinxi Lyu, Ari Holtzman, Mikel Artetxe, Mike Lewis, Hannaneh Hajishirzi, and Luke Zettlemoyer. 2022.
\newblock \href {https://doi.org/10.18653/v1/2022.emnlp-main.759} {Rethinking the role of demonstrations: What makes in-context learning work?}
\newblock In \emph{Proceedings of the 2022 Conference on Empirical Methods in Natural Language Processing}, pages 11048--11064, Abu Dhabi, United Arab Emirates. Association for Computational Linguistics.

\bibitem[{OpenAI(2023{\natexlab{a}})}]{achiam2023gpt}
OpenAI. 2023{\natexlab{a}}.
\newblock Gpt-4 technical report.
\newblock \emph{arXiv preprint arXiv:2303.08774}.

\bibitem[{OpenAI(2023{\natexlab{b}})}]{openai2023gpt4}
OpenAI. 2023{\natexlab{b}}.
\newblock \href {https://arxiv.org/abs/2303.08774} {Gpt-4 technical report}.
\newblock \emph{arXiv preprint arXiv:2303.08774}.

\bibitem[{Ouyang et~al.(2022)Ouyang, Wu, Jiang, Almeida, Wainwright, Mishkin, Zhang, Agarwal, Slama, Ray et~al.}]{ouyang2022training}
Long Ouyang, Jeff Wu, Xu~Jiang, Diogo Almeida, Carroll~L Wainwright, Pamela Mishkin, Chong Zhang, Sandhini Agarwal, Katarina Slama, Alex Ray, et~al. 2022.
\newblock \href {https://arxiv.org/abs/2203.02155} {Training language models to follow instructions with human feedback}.
\newblock \emph{arXiv preprint arXiv:2203.02155}.

\bibitem[{Radford et~al.(2019)Radford, Wu, Child, Luan, Amodei, Sutskever et~al.}]{radford2019language}
Alec Radford, Jeffrey Wu, Rewon Child, David Luan, Dario Amodei, Ilya Sutskever, et~al. 2019.
\newblock \href {https://cdn.openai.com/better-language-models/language_models_are_unsupervised_multitask_learners.pdf} {Language models are unsupervised multitask learners}.
\newblock \emph{OpenAI blog}, 1(8):9.

\bibitem[{Raffel et~al.(2020)Raffel, Shazeer, Roberts, Lee, Narang, Matena, Zhou, Li, Liu et~al.}]{raffel2020exploring}
Colin Raffel, Noam Shazeer, Adam Roberts, Katherine Lee, Sharan Narang, Michael Matena, Yanqi Zhou, Wei Li, Peter~J Liu, et~al. 2020.
\newblock \href {https://www.jmlr.org/papers/volume21/20-074/20-074.pdf?ref=https://githubhelp.com} {Exploring the limits of transfer learning with a unified text-to-text transformer.}
\newblock \emph{J. Mach. Learn. Res.}, 21(140):1--67.

\bibitem[{Reimers and Gurevych(2019)}]{reimers-gurevych-2019-sentence}
Nils Reimers and Iryna Gurevych. 2019.
\newblock \href {https://doi.org/10.18653/v1/D19-1410} {Sentence-{BERT}: Sentence embeddings using {S}iamese {BERT}-networks}.
\newblock In \emph{Proceedings of the 2019 Conference on Empirical Methods in Natural Language Processing and the 9th International Joint Conference on Natural Language Processing (EMNLP-IJCNLP)}, pages 3982--3992, Hong Kong, China. Association for Computational Linguistics.

\bibitem[{Roberts et~al.(2020)Roberts, Raffel, and Shazeer}]{roberts-etal-2020-much}
Adam Roberts, Colin Raffel, and Noam Shazeer. 2020.
\newblock \href {https://doi.org/10.18653/v1/2020.emnlp-main.437} {How much knowledge can you pack into the parameters of a language model?}
\newblock In \emph{Proceedings of the 2020 Conference on Empirical Methods in Natural Language Processing (EMNLP)}, pages 5418--5426, Online. Association for Computational Linguistics.

\bibitem[{Rubin et~al.(2022)Rubin, Herzig, and Berant}]{rubin-etal-2022-learning}
Ohad Rubin, Jonathan Herzig, and Jonathan Berant. 2022.
\newblock \href {https://doi.org/10.18653/v1/2022.naacl-main.191} {Learning to retrieve prompts for in-context learning}.
\newblock In \emph{Proceedings of the 2022 Conference of the North American Chapter of the Association for Computational Linguistics: Human Language Technologies}, pages 2655--2671, Seattle, United States. Association for Computational Linguistics.

\bibitem[{Sun et~al.(2022)Sun, Wang, Tay, Yang, and Zhou}]{sun2022recitation}
Zhiqing Sun, Xuezhi Wang, Yi~Tay, Yiming Yang, and Denny Zhou. 2022.
\newblock \href {https://arxiv.org/abs/2210.01296} {Recitation-augmented language models}.
\newblock \emph{arXiv preprint arXiv:2210.01296}.

\bibitem[{Taori et~al.(2023)Taori, Gulrajani, Zhang, Dubois, Li, Guestrin, Liang, and Hashimoto}]{alpaca}
Rohan Taori, Ishaan Gulrajani, Tianyi Zhang, Yann Dubois, Xuechen Li, Carlos Guestrin, Percy Liang, and Tatsunori~B. Hashimoto. 2023.
\newblock Stanford alpaca: An instruction-following llama model.
\newblock \url{https://github.com/tatsu-lab/stanford_alpaca}.

\bibitem[{Voorhees et~al.(1999)}]{voorhees1999trec}
Ellen~M Voorhees et~al. 1999.
\newblock \href {https://trec.nist.gov/pubs/trec8/papers/qa_report.pdf} {The trec-8 question answering track report.}
\newblock In \emph{Trec}, volume~99, pages 77--82.

\bibitem[{Wang et~al.(2022)Wang, Srikumar, Hajishirzi, and Smith}]{wang2022elaboration}
Wenya Wang, Vivek Srikumar, Hanna Hajishirzi, and Noah~A Smith. 2022.
\newblock \href {https://arxiv.org/abs/2209.01232} {Elaboration-generating commonsense question answering at scale}.
\newblock \emph{arXiv preprint arXiv:2209.01232}.

\bibitem[{Wei et~al.(2022{\natexlab{a}})Wei, Bosma, Zhao, Guu, Yu, Lester, Du, Dai, and Le}]{wei2022finetuned}
Jason Wei, Maarten Bosma, Vincent Zhao, Kelvin Guu, Adams~Wei Yu, Brian Lester, Nan Du, Andrew~M. Dai, and Quoc~V Le. 2022{\natexlab{a}}.
\newblock \href {https://openreview.net/forum?id=gEZrGCozdqR} {Finetuned language models are zero-shot learners}.
\newblock In \emph{International Conference on Learning Representations}.

\bibitem[{Wei et~al.(2022{\natexlab{b}})Wei, Wang, Schuurmans, Bosma, Chi, Le, and Zhou}]{wei2022chain}
Jason Wei, Xuezhi Wang, Dale Schuurmans, Maarten Bosma, Ed~Chi, Quoc Le, and Denny Zhou. 2022{\natexlab{b}}.
\newblock \href {https://arxiv.org/abs/2201.11903} {Chain of thought prompting elicits reasoning in large language models}.
\newblock \emph{arXiv preprint arXiv:2201.11903}.

\bibitem[{Ye et~al.(2022)Ye, Gao, Li, Xu, Feng, Wu, Yu, and Kong}]{ye2022zerogen}
Jiacheng Ye, Jiahui Gao, Qintong Li, Hang Xu, Jiangtao Feng, Zhiyong Wu, Tao Yu, and Lingpeng Kong. 2022.
\newblock \href {https://arxiv.org/abs/2202.07922} {Zerogen: Efficient zero-shot learning via dataset generation}.
\newblock \emph{arXiv preprint arXiv:2202.07922}.

\bibitem[{Yu et~al.(2022)Yu, Iter, Wang, Xu, Ju, Sanyal, Zhu, Zeng, and Jiang}]{yu2022generate}
Wenhao Yu, Dan Iter, Shuohang Wang, Yichong Xu, Mingxuan Ju, Soumya Sanyal, Chenguang Zhu, Michael Zeng, and Meng Jiang. 2022.
\newblock \href {https://arxiv.org/abs/2209.10063} {Generate rather than retrieve: Large language models are strong context generators}.
\newblock \emph{arXiv preprint arXiv:2209.10063}.

\bibitem[{Zhang et~al.(2022{\natexlab{a}})Zhang, Chen, Xu, Cao, Chen, Cohn, and Fang}]{zhang2022survey}
Qin Zhang, Shangsi Chen, Dongkuan Xu, Qingqing Cao, Xiaojun Chen, Trevor Cohn, and Meng Fang. 2022{\natexlab{a}}.
\newblock \href {https://arxiv.org/abs/2211.07886} {A survey for efficient open domain question answering}.
\newblock \emph{arXiv preprint arXiv:2211.07886}.

\bibitem[{Zhang et~al.(2023)Zhang, Chen, Xu, Cao, Chen, Cohn, and Fang}]{zhang-etal-2023-survey-efficient}
Qin Zhang, Shangsi Chen, Dongkuan Xu, Qingqing Cao, Xiaojun Chen, Trevor Cohn, and Meng Fang. 2023.
\newblock \href {https://doi.org/10.18653/v1/2023.acl-long.808} {A survey for efficient open domain question answering}.
\newblock In \emph{Proceedings of the 61st Annual Meeting of the Association for Computational Linguistics (Volume 1: Long Papers)}, pages 14447--14465, Toronto, Canada. Association for Computational Linguistics.

\bibitem[{Zhang et~al.(2022{\natexlab{b}})Zhang, Roller, Goyal, Artetxe, Chen, Chen, Dewan, Diab, Li, Lin et~al.}]{zhang2022opt}
Susan Zhang, Stephen Roller, Naman Goyal, Mikel Artetxe, Moya Chen, Shuohui Chen, Christopher Dewan, Mona Diab, Xian Li, Xi~Victoria Lin, et~al. 2022{\natexlab{b}}.
\newblock \href {https://arxiv.org/abs/2205.01068} {Opt: Open pre-trained transformer language models}.
\newblock \emph{arXiv preprint arXiv:2205.01068}.

\bibitem[{Zhang et~al.(2022{\natexlab{c}})Zhang, Zhang, Li, and Smola}]{zhang2022automatic}
Zhuosheng Zhang, Aston Zhang, Mu~Li, and Alex Smola. 2022{\natexlab{c}}.
\newblock \href {https://arxiv.org/abs/2210.03493} {Automatic chain of thought prompting in large language models}.
\newblock \emph{arXiv preprint arXiv:2210.03493}.

\bibitem[{Zhao et~al.(2021)Zhao, Wallace, Feng, Klein, and Singh}]{zhao2021calibrate}
Zihao Zhao, Eric Wallace, Shi Feng, Dan Klein, and Sameer Singh. 2021.
\newblock \href {http://proceedings.mlr.press/v139/zhao21c.html} {Calibrate before use: Improving few-shot performance of language models}.
\newblock In \emph{International Conference on Machine Learning}, pages 12697--12706. PMLR.

\bibitem[{Zhu et~al.(2021)Zhu, Lei, Wang, Zheng, Poria, and Chua}]{zhu2021retrieving}
Fengbin Zhu, Wenqiang Lei, Chao Wang, Jianming Zheng, Soujanya Poria, and Tat-Seng Chua. 2021.
\newblock \href {https://arxiv.org/abs/2101.00774} {Retrieving and reading: A comprehensive survey on open-domain question answering}.
\newblock \emph{arXiv preprint arXiv:2101.00774}.

\end{thebibliography}

\clearpage
\appendix

\section{Statistics of the datasets}
\label{app:testset_stat}
The statistics of the datasets we use in Table \ref{tab:data_stat}. We find that test samples in TriviaQA have much longer question length than the other two datasets. TriviaQA also provides more reference answer, so models tend to have a higher score on this dataset.

\begin{table}
  \centering\small
  \setlength{\tabcolsep}{6pt}
    \begin{tabular}{lccccc}
    \toprule
    Datasets & Size & Q Words & A Words & Answers \\
    \midrule
    WebQ  & 2.0K & 6.8   & 2.5 & 2.4 \\
    NQ    & 3.6K & 9.1  & 2.2  & 2.0 \\
    TriviaQA & 11K & 14.0 & 2.5 & 14.0 \\
    \bottomrule
    \end{tabular}%
  \caption{Statistics for the test split of each dataset. Answers, Q Words and A Words refer to the average number reference answers, words in question, and words in answer for each sample respectively.}
  \label{tab:data_stat}%
\end{table}%

\section{Names and Required Number of Examples for Topics}
\label{app:gen_topic}
We list the name and required number of examples for each topic in Table \ref{tab:topic_name_number}. In general, we require about 100 examples for each high-level category (e.g., \textit{Art, History, Sports}).

\section{Details in Generating Pseudo QA Dataset}
\label{app:gen_details}
We present the exact templates and API parameters used in self-generation steps in Table \ref{tab:gen_details}. After getting the passage, we use the NLTK toolkit\footnote{https://www.nltk.org} to remove the last sentence of it if the last sentence is incomplete or truncated. 

\section{Statistics of the Generated Pseudo QA datasets}
We show some statistics of the generated pseudo QA dataset in Table \ref{tab:pseudo_stat}.

\section{Specific Templates in Different Formats}
\label{app:format_template}
The specific templates used in main experiments (QAE) and in Sec \ref{sec:format} are shown in Table \ref{tab:format_template}. We refer to \citet{yu2022generate} to design them. The \texttt{temperature} is 0 for all formats.

\begin{table}
  \centering
  \caption{Names and required examples for the 29 manually designed topics.}
    \begin{tabular}{lc}
    \toprule
    Topic & Number \\
    \midrule
    politician & 100 \\
    athlete & 40 \\
    sports team & 40 \\
    sports event & 40 \\
    country & 40 \\
    city  & 60 \\
    historical figure & 50 \\
    historical event & 50 \\
    war   & 40 \\
    religion & 20 \\
    singer & 50 \\
    song  & 50 \\
    actor or actress & 50 \\
    movie or TV series & 50 \\
    writer & 30 \\
    book  & 30 \\
    painter & 30 \\
    painting & 30 \\
    composer & 30 \\
    classical music & 30 \\
    tourist attraction & 100 \\
    scientist & 40 \\
    scientific term & 40 \\
    video game & 40 \\
    animal & 40 \\
    plant & 40 \\
    food  & 40 \\
    enterprise & 50 \\
    international organization & 50 \\
    \bottomrule
    \end{tabular}%
  \label{tab:topic_name_number}%
\end{table}%

\begin{table}
  \centering
  \caption{Statistics of the generated pseudo
QA dataset.}
  \setlength{\tabcolsep}{20pt}
    \begin{tabular}{lr}
    \toprule
    \textbf{Avg. Words} &  \\
    \midrule
    Passage & 77.41 \\
    Question & 9.52 \\
    Answer & 2.04 \\
    Explanation & 13.37 \\
    \midrule
    \textbf{Question Type} &  \\
    \midrule
    where & 204 \\
    how   & 335 \\
    who   & 898 \\
    what  & 2,814 \\
    when  & 475 \\
    which & 151 \\
    others & 6 \\
    \bottomrule
    \end{tabular}%
  \label{tab:pseudo_stat}%
\end{table}%

\begin{table*}
  \centering
  \caption{The indexing and searching time for Self-Prompting.}
    \setlength{\tabcolsep}{4.5pt}
    \begin{tabular}{l|ll|ll}
    \toprule
    Device & \multicolumn{2}{l|}{Indexing}       & \multicolumn{2}{l}{Search time (1,000 queries)}  \\
\cmidrule{2-5}          & Encoding pseudo data & Clustering & Encoding queries& Cluster-based Retrieval \\
    \midrule
    CPU   & 25.33s & 1.86s & 4.20s & 3.21s \\
    1 GPU & 5.82s & 1.69s & 0.30s & 1.57s \\
    \bottomrule
    \end{tabular}%
  \label{tab:running_efficiency}%
\end{table*}%

\section{Training Small Models with the Pseudo QA dataset}
\label{app: use-pseudo-data-to-train}
We investigate if the built pseudo ODQA dataset can be used to train a specialized smaller model. To implement it, we retrain the reader of a RAG \citep{lewis2020retrieval} pre-trained on the full NQ trining set\footnote{https://huggingface.co/facebook/rag-sequence-nq} on our pseudo data from BART-large, while fixing the retriever module (we are unable to jointly train both modules as the original RAG did with our synthetic data). The EM and F1 on the NQ subset used in the analysis section are in Table \ref{tab:retraining_rag}.

This huge performance gap is likely caused by: 1) The pre-trained model benefits from more in-domain examples (70K+ NQ training examples vs 5K pseudo data). There is also significant train-test overlap in NQ as reported in \citet{lewis-etal-2021-question}, 2) Retraining only the reader while fixing the retriever causes a mismatch, and 3) The reader is trained to only process one passage at a time, instead of comparing multiple passages simultaneously as the pre-trained reader. Given these results, we conclude that in current setting, it is a better practice to use the generated data for in-context learning samples of LLM itself than training a smaller supervised model.


\begin{table}
  \centering\small
  \caption{The performance of two different RAG models on the NQ subset.}
    \begin{tabular}{l|cc}
    \toprule
    Model & EM & F1 \\
    \midrule
    Pre-trained RAG & 36.10  & 44.01 \\
    RAG, reader retrained & 19.00    & 27.44 \\
    \bottomrule
    \end{tabular}%
  \label{tab:retraining_rag}%
\end{table}%

\begin{table}
  \centering\small
  \caption{Comparison between Self-Prompting and a real-time generation variant.}
    \begin{tabular}{l|cc}
    \toprule
    Method & EM    & F1 \\
    \midrule
    Self-Prompting & 37.2  & 47.6 \\
    Real-time Generation Variant & 35.3  & 45.5 \\
    \bottomrule
    \end{tabular}%
  \label{tab:real-time-selfp}%
\end{table}%


\section{Real-time Pseudo Data Generation Variant for Self-Prompting}
\label{app:real-time-selfp}
We also investigate the potential of a real-time generation variant of Self-Prompting. Specifically, for each test sample, we first ask the LLM to identify the main entity in the question and then generate a short passage based on it. The rest generation steps are all kept the same. In short, we reinforce the relevance of in-context demonstrations but weaken the diversity. We show the performance of this variant on the NQ subset in the analysis section. 

The results in Table \ref{tab:real-time-selfp} show that this real-time variant does not bring improvement over Self-Prompting. This aligns with our earlier finding that cluster-based selection outperforms global similarity retrieval (in Table \ref{tab:selection_ana}), indicating diversity is important for effective prompting. 
While real-time generation of highly-related examples is promising intuitively, it seems that some variability in the prompts is beneficial to avoid overfitting to a narrow context. 
Additionally, the higher inference delay and cost of real-time generation make it less practical. 
Overall, these results emphasize the balance of relevance and diversity when constructing demonstrations, and some offline preparation can be more efficient.

\section{Cost and Running Efficiency}
\label{app:cost-running-efficiency}
We provide a brief analysis of the cost and running efficiency of Self-Prompting. The total cost for constructing the pseudo QA dataset is about \$120 and it takes about 6 hours to collect the data.

The indexing and search time over the pseudo data in Table \ref{tab:running_efficiency}. As a comparison, the encoding/indexing/search time for DPR \citep{karpukhin-etal-2020-dense} (statistics from their original paper) is 8.8h for encoding the complete Wikipedia dump on an 8-GPU machine, 8.5h for indexing the Wikipedia passage embeddings, and the search rate is 995 queries per second. 

\section{More Data Quality Analysis}
\begin{figure*}[htbp]
    \centering
    \subfigbottomskip=0pt
    \subfigure[Same to Figure \ref{fig:quantitative_eval_data}(a), but normalized by passage length.]{
        \begin{minipage}[t]{0.45\linewidth}
            \centering
            \includegraphics[width=0.95\textwidth]{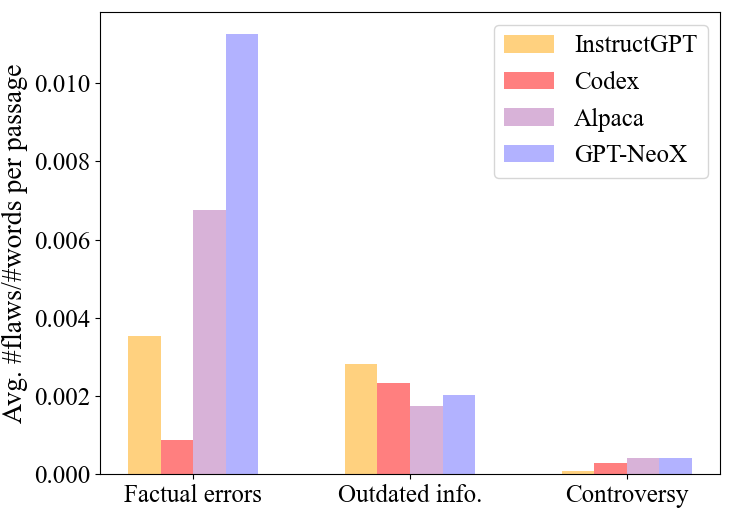}
        \end{minipage}
    }
    \subfigure[Ratio of passages with specific kind of flaws.]{
        \begin{minipage}[t]{0.45\linewidth}
            \centering
            \includegraphics[width=0.95\textwidth]{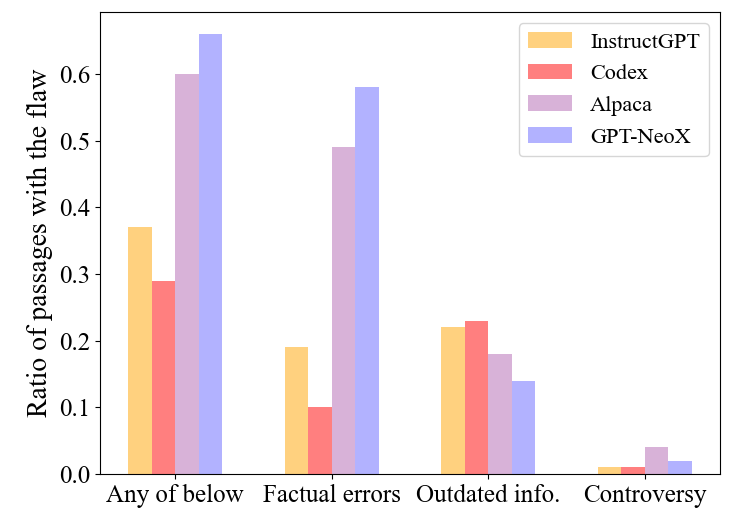}
        \end{minipage}
    }
    \caption{More statistics of quality analysis for our generated pseudo data.}
    \label{fig:quantitative_eval_data_more}
\end{figure*}

We provide more statistics for the quality analysis on our generated pseudo data in Figure \ref{fig:quantitative_eval_data_more} as a supplement to Sec \ref{sec: data_quality}, including the average number of flaws in the generated passages normalized by passage length, and the ratio of passages with at least one flaw. Similar to Figure \ref{fig:quantitative_eval_data}, they also show the superiority of the OpenAI models over Alpaca and GPT-NeoX.

\begin{table*}
  \centering\small
  \caption{Parameters and detailed prompts used in each generation step, the temperatures are all set as 0 (except generating diverse examples for a certain topic) to increase the correctness of the generated content at each step.}
    \begin{tabular}{l|p{8cm}|c|c}
    \toprule
    Generation Step & Prompt & max\_tokens & temperature \\
    \midrule
    \midrule
    Example & List some \{topic\}, separated by '|': & 1024  & 1 \\
    \midrule
    Passage & This is a passage from Wikipedia about the \{topic\}, \{example\}:\blue{\textbackslash n} & 256   & 0 \\
    \midrule
    Entity & Here is a passage: \{passage\}\blue{\textbackslash n\textbackslash n}Extract the named entities (like date, location, organization, character, number) in it, and separate them by '|'. If no named entity in it, write 'None' only. & 50    & 0 \\
    \midrule
    Question & \{passage\} \blue{\textbackslash n}\{entity\} is the answer to the question: & 50    & 0 \\
    \midrule
    Double Check & Passage: \{passage\}\blue{\textbackslash n}Question: \{question\}\blue{\textbackslash n}Short Answer (extracted from the passage, less than 6 words): & 50    & 0 \\
    \midrule
    Explain & Passage: \{passage\}\blue{\textbackslash n}Question: \{question\} Answer: \{answer\}\blue{\textbackslash n}You can refer to the passage and write a short explanation to this Question-Answer pair, "\{answer\}" must in the explanation: & 50    & 0 \\
    \bottomrule
    \end{tabular}%
  \label{tab:gen_details}%
\end{table*}%

\begin{table*}
  \centering\small
  \caption{Specific templates for different input formats and the parameters of calling APIs.}
    \begin{tabular}{l|p{11cm}|c}
    \toprule
    Format & Template & max\_tokens \\
    \midrule
    \midrule
    \multicolumn{3}{l}{\textit{*one iteration}} \\
    QAE   & Question: \{question\} \blue{\textbackslash n\textbackslash n} The answer (just one entity) is \{answer\} because \{explanation\} & 128 \\
    QAP   & Question: \{question\} \blue{\textbackslash n\textbackslash n} The answer (just one entity) is \{answer\} referring to the passage: \{passage\} & 256 \\
    QEA   & Question: \{question\} \blue{\textbackslash n\textbackslash n} \{explanation\} So the answer (just one entity) is \{answer\} & 256 \\
    QPA   & Question: \{question\} \blue{\textbackslash n\textbackslash n} \{passage\} So the answer (just one entity) is \{answer\} & 256 \\
    QAEP  & Question: \{question\} \blue{\textbackslash n\textbackslash n} The answer (just one entity) is \{answer\} because \{explanation\} referring to the passage: \{passage\} & 256 \\
    QAPE  & Question: \{question\} \blue{\textbackslash n\textbackslash n} The answer (just one entity) is \{answer\} referring to the passage: \{passage\} so \{explanation\} & 256 \\
    \midrule
    \midrule
    \multicolumn{3}{l}{\textit{*two iterations}} \\
    1. QP & Generate a background document from Wikipedia to answer the given question. \{question\}\textbackslash{}n\{passage\}  & 256 \\
    2. PQA & Passage: \{passage\} \blue{\textbackslash n\textbackslash n} Question: \{question\} \blue{\textbackslash n\textbackslash n} Referring to the passage above, the correct answer (just one entity) to the given question is \{answer\} & 20 \\
    \midrule
    1. QE & Generate a piece of evidence to answer the given question. \{question\}\textbackslash{}n\{explanation\} & 256 \\
    2. EQA & Evidence: \{explanation\} \blue{\textbackslash n\textbackslash n} Question: \{question\} \blue{\textbackslash n\textbackslash n} Referring to the evidence above, the correct answer (just one entity) to the given question is \{answer\} & 20 \\
    \bottomrule
    \end{tabular}%
  \label{tab:format_template}%
\end{table*}%


\end{document}